# Large-Scale Multipurpose Benchmark Datasets For Assessing Data-Driven Deep Learning Approaches For Water Distribution Networks [†]

Andrés Tello [1,*], Huy Truong [1,*], Alexander Lazovik[1] and Victoria Degeler [2]

1. Bernoulli Institute, University of Groningen, Groningen, The Netherlands
2. Informatics Institute, University of Amsterdam, Amsterdam, The Netherlands
* Both authors contributed equally. Correspondence: andres.tello@rug.nl, h.c.truong@rug.nl
† Presented at WDSA CCWI, Ferrara, Italy, July 01st – 04th.

**Abstract:** Currently, the number of common benchmark datasets that researchers can use straight away for assessing data-driven deep learning approaches is very limited. Most studies provide data as configuration files. It is still up to each practitioner to follow a particular data generation method and run computationally intensive simulations to obtain usable data for model training and evaluation. In this work, we provide a collection of datasets that includes several small and medium size publicly available Water Distribution Networks (WDNs), including Anytown, Modena, Balerma, C-Town, D-Town, L-Town, Ky1, Ky6, Ky8, Ky10, and Ky13. In total 1,394,400 hours of WDNs data operating under normal conditions is made available to the community.

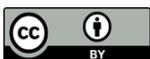

## 1. Introduction

Researchers and practitioners working in Water Distribution Networks (WDNs) Analysis need to face the challenges associated with data availability. Privacy, safety and costs are the main cause of the aforementioned data limitation [1][2]. The earliest benchmark datasets aimed to facilitate the planning, design, and management of WDNs. Such datasets included the WDN topologies and the physical properties of the systems presented as configuration files. Those files serve as input to physics-based models that leverage mathematical tools to simulate the system's hydraulics. Anytown [3], is one example among the first data contributions to the water research community. Later, the well-known C-Town dataset [4] was used in the "Battle of Water Calibration Networks". Another important contribution in terms of WDNs data is the Kentucky Research Database [5]. It includes 12 small and medium real WDNs in the state of Kentucky in the United States. These approaches provide a collection of configuration files that can be used for generating WDNs data, but not the data itself. The only work known which provides actual data is LeakDB [6], but the WDNs used in that work are very small (Net1: 10 nodes and Hanoi: 32 nodes). Moreover, the search space of the input parameters used in LeakDB is reduced to demand, and pipe rough coefficients, diameter and length. Thus, limiting the variability among the generated scenarios.

Recently, researchers in the water domain have turned into data-driven methods when performing WDNs analysis. Several deep learning approaches have been used to solve a variety of problems, e.g., estate estimation, leakage detection and localization, water demand forecasting, among others. Nonetheless, all those methods are known to be data-hungry, i.e., they require vast amounts of data for model training. Those data requirements for training deep learning models expose the shortcomings associated with the existing datasets. First, the number of common benchmark datasets that can be used straight away for multi-task purposes is very limited. Second, existing data rarely include time-dependent patterns, or few patterns are overused by being assigned multiple times to different nodes.

Although the previously mentioned data files are being successfully used for different optimization problems, they do not contain readily usable data. Then, most of the so-called "datasets" are a collection of configuration files that can be used as input for mathematical modeling tools, like EPANET, in order to generate usable data. They do not contain pressures, flow rates, velocities, etc. All these data, required for different analysis purposes, need to be obtained by means of mathematical simulations. Moreover, those configuration files represent a single instance of the network state determined by the configuration values. Then, in order to create multiple WDN states, the configuration files need to be manually adjusted and the simulation needs to be executed several times with the new inputs. Other works proposed algorithms for data generation [7, 8], which automate the simulation executions. While partially solving the problem, those approaches do not provide data itself. It is still up to each researcher or practitioner to follow a particular data generation method from the literature and adjust it using their own rationale before obtaining usable data.

Temporal-dependent patterns are required for the simulations to model the dynamic input parameters, e.g., demand or pump curve. Those patterns are rarely provided in publicly available data files due to privacy and safety concerns. At best, the available data only include a few time-dependent patterns which are reused multiple times for different nodes in the network. Such overuse produces scenarios where multiple nodes resemble exactly the same water consumption behavior, which is unrealistic [8]. This approach limits the variety of the data and the robustness of the models to the uncertainties encountered in real-world cases.

In this work, we attempt to fill the data availability gap by providing a collection of datasets that includes several publicly available Water Distribution Networks. In the current version, our dataset contains small and medium size WDNs including Anytown, Modena, Balerma, C-Town, D-Town, L-Town, Ky1, Ky6, Ky8, Ky10, and Ky13. Following a modified version of the approach presented in [8], we generated ready-to-use data that represents stable states of 10 WDNs operating under normal conditions. In addition, we propose a demand patterns generation method which allows us to assign a different 24-hour time-series demand for each individual node for all networks. As a result, 1,394,400 WDN states are provided. In addition, we provide a small-version of the data that comprises 1,000 scenarios per network with a total of 240,000 hours of simulated data. The data provided enables researchers to address the following tasks: (i) State Estimation, leveraging a limited number of sensors to reconstruct target measurements such as pressure, flow, and head; (ii) Demand Forecasting, with the goal of predicting the customer demand within a given period; (iii) Surrogate Modeling efficiently replicating the behavior of a simulation.

The remainder of this paper is organized as follows. Section 2 presents our dataset, describing the parameters selection, parameters boundary determination and data encapsulation for data sharing. Section 3 describes the most important characteristics of our dataset and how they differ from existing data benchmarks. Finally, in Section 4, we present the conclusions of our work.

## 2. Dataset creation

In pursuit of enhanced generalization, the dataset extends the conventional simulation's capability to untouched hydraulic-related parameters. The simulation is wrapped by a utility tool so-called WNTR [9]. The data generation method involves three phases: Parameter selection, parameter boundary and sample quantity identification, and data encapsulation.

**Parameter Selection.** We collected several public WDNs from diverse regions. For each WDN, its relevant information (e.g., network topology, nodal elevation, and demands, etc), is compressed into an input (INP) file. We initiated by extracting all hydraulic parameters associated with every existing component, encompassing reservoirs, tanks, junctions, pipes, pumps, and various types of valves. It is worth noting that general information (name, coordinates, etc…) and quality-related parameters were omitted due to the inconsideration in the scope of this study and the optimal storage matter. Afterward, a filtering approach alleviated irrelevant and redundant parameters from the consideration. For example, a demand parameter was represented by base demand—a scaling scalar, and multipliers—a vector representing the demand pattern. In the downstream applications, keeping both representations was unnecessary, resulting in only storing calculated demand values in the first-encountered parameter. In total, sixty hydraulic parameters were considered, and, after the filtering, half of them were ready to use.

**Parameter boundary and sample quantity identification.** The next phase included the following steps: parameter boundary determination and sample quantity selection w.r.t WDN. The spectrum of input parameters in the pre-simulation stage is crucial for defining the data space. In essence, we utilized random sampling, which is a simple yet robust generation strategy, to synthetically generate numerous scenarios within the simulation. In light of this, the broader the parameter range, the more extensive the diversity.

In terms of demand, we automatically generated demand patterns by defining a consumption profile. The profile defined very-low, low, mid, and high water consumption ranges. First, the day was divided into four 6-hour segments starting at midnight. Then, the consumption ranges were randomly applied to the segments of a day, and using a periodic function we extend these randomly generated patterns along the time axis to reach a specified duration.

To restrict outlier scenarios created by a corrupted set of parameters, we enforced several rules of validation. In particular, we targeted the pressure, one of the simulation outputs, and claimed its values in a range of [0, 151] [8]. Nevertheless, the restriction posed a significant challenge in finding optimal parameter boundaries for each WDN due to the massive search space. With an anticipated 100 scenarios, an arbitrary selection for each parameter frequently resulted in zero successful outcomes. As such, we designed a semi-automatic algorithm searching for an optimal configuration. For every parameter, the algorithm restricted values within a global data space crafted from available WDNs, then strategically selected discrete points along the max-min line. The criterion for evaluating the "goodness" of selected values was the ratio of successful attempts out of 100 cases. By default, this acceptable ratio was set within the range of 40% to 60%. Noting that after multiple attempts without satisfying the essential ratio, the algorithm preserved the original value (if present) from the input file. This approach empowers practitioners to manually refine these parameters, fostering a further study to discover the optimal configuration for each WDN.

Another consideration was the amount of created scenarios for each WDN. Each network varies in number of nodes, edges, and the topological complexity. As such, creating the same amount of samples for all WDNs could increase the burden of computational and storage units. Nonetheless, there existed no general rule for dataset size estimation. With this in mind, we empirically adopted a rule of 10, generating records in a quantity 10 times greater than the number of nodes within a WDN. The detailed amount and relevant information can be seen in Table 1.

**Table 1.** Overview of Water Distribution Networks for sample generation.

| WDN | Junctions | Pipes | Demand ($10^3 \times m^3/s$)* | Pressure (m)* | Nr. Scenarios |
|---|---|---|---|---|---|
| Anytown | 19 | 40 | [2.7, 24.3] | [26.70, 62.33] | 190 |
| Balerma | 443 | 454 | [0.5, 3.00] | [38.00, 66.96] | 4,430 |
| C-Town | 388 | 429 | [0.2, 1.2] | [39.75, 69.98] | 3,880 |
| D-town | 399 | 443 | [0.2, 0.9] | [42.31, 78.50] | 3,990 |
| Ky1 | 856 | 984 | [0.3, 1.2] | [46.28, 95.28] | 8,560 |
| Ky6 | 543 | 644 | [0.1, 0.8] | [42.16, 90.37] | 5,430 |
| Ky8 | 1,325 | 1,614 | [0.06, 0.4] | [56.47, 111.03] | 13,250 |
| Ky13 | 778 | 940 | [0.4, 2.1] | [33.57, 51.63] | 7,780 |
| L-Town | 782 | 905 | [0.06, 0.4] | [1.50, 54.22] | 7,820 |
| Modena | 268 | 317 | [0.5, 2.9] | [25.89, 35.16] | 2,680 |

(*) Demand and pressures are expressed in the [Q1, Q3] range.

**Data encapsulation.** In the last step, we generated a random matrix extracted from the optimal configuration and utilized it in a large-scale simulation. Specifically, we leveraged a cluster of 32 CPUs to run simulations in parallel, each with a batch size of 64. Notably, the longest run endured for approximately 4 uninterrupted days. Afterward, both input parameters and output simulation data were encoded and encapsulated by Zarr, a robust Python library renowned for handling chunked, compressed, and labeled arrays efficiently and at scale. In total, the dataset has occupied more than 50 gigabytes of compressed storage space. Furthermore, we strategically created 1,000 scenarios per network in CSV format for serving a broader audience. This compact dataset has been assigned a DOI and can be accessed through the Data Availability Statement section.

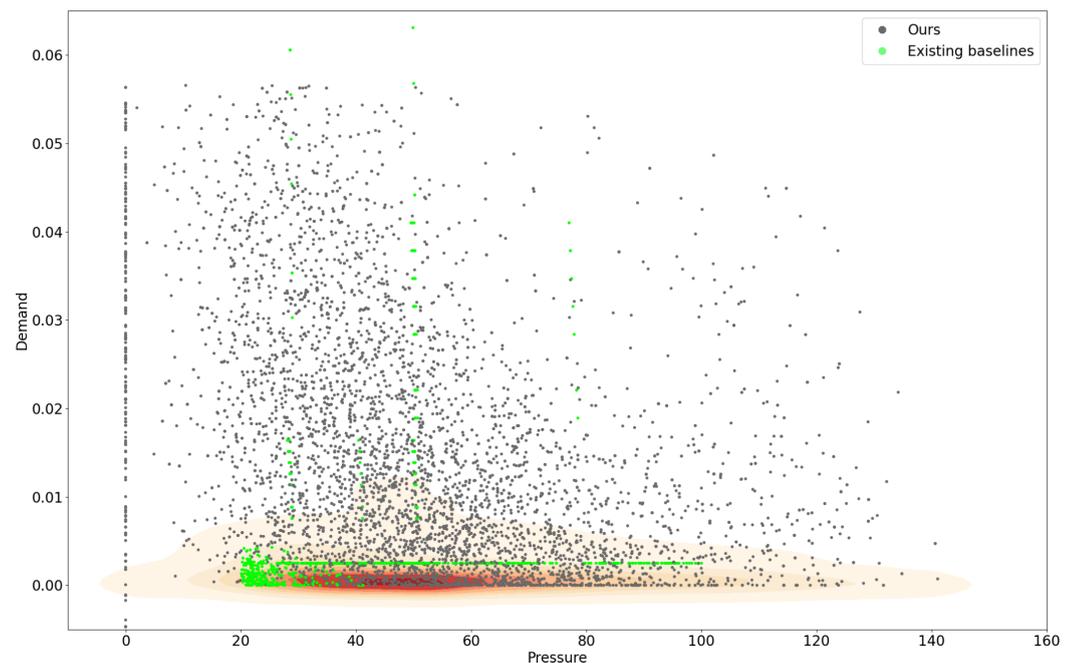

**Figure 1.** Density distribution of pressure and demand of our generated data (gray) vs the WDNs baselines (green).

## 3. Discussion

In this section, we dive into the characteristics of the dataset provided. Figure 1 showcases the dispersion of pressure and demand data points between the baseline WDNs and the synthetic sets generated by our method. Within a valid pressure range, the baseline WDNs exhibit a concentration of points around 20m, caused by the small values of demand (close to zero) in baselines. Such a situation can lead to instability in practical scenarios. While the baselines cover pressures from 20 to 100m, our pressure data expands a higher range, covering pressures up to 140m. From Figure 1 it is clear that our dataset covers a much higher spectrum in terms of demand, contributing to the variety of the data. This variability is essential for training cutting-edge data-driven algorithms, providing adaptability and insight into diverse real-world scenarios.

## 4. Conclusion

This study highlighted the limited amount of ready-to-use data for WDNs Analysis, showing that existing data collections comprise mostly input configuration files for mathematical simulations. Researchers still need to run computationally intensive simulations and tweak existing data generation algorithms in order to obtain usable data. We provide a dataset of 50 gigabytes of compressed data which includes 1,394,400 WDNs states operating under normal conditions. The data include all the input parameters used for data generation, and the outputs obtained with the simulation tool WNTR, e.g., pressure, flow rate, velocity, head, etc. This work is, to the best of our knowledge, the first large-scale dataset that contains that can be downloaded and used right away for different WDNs Analyses.

**Funding:** This work is funded by the project DiTEC: Digital Twin for Evolutionary Changes in Water Networks (NWO 19454), and by NWO C2D and TKI HTSM Ecida Project Grant No. 628011003.

**Data Availability Statement:** The compact dataset associated with this study has been found in the DOI link (https://doi.org/10.5281/zenodo.10974087).

Given the external quota limit, we have made the larger dataset accessible through: https://drive.google.com/drive/folders/1nPO7qfOBAoUSRLozyqfaf7hxrBnFV-7g

**Acknowledgments:** We thank the Center for Information Technology of the University of Groningen for their support and for providing access to the Hábrók high performance computing cluster.